\documentclass[letterpaper, 10 pt, conference]{ieeeconf}  

\IEEEoverridecommandlockouts                              

\usepackage{graphics} 
\usepackage{epsfig} 
\usepackage{times} 
\usepackage{amsmath} 
\usepackage{amssymb}  
\usepackage{hyperref}
\usepackage{multirow}
\usepackage{caption,setspace}
\usepackage{booktabs}
\usepackage{arydshln}
\usepackage{tablefootnote}
\usepackage{csquotes}
\usepackage{footnote}
\makesavenoteenv{tabular}
\makesavenoteenv{table}
\usepackage[]{threeparttable}
\usepackage{breqn}
\usepackage{subcaption}
\usepackage{algorithm}
\usepackage[noend]{algpseudocode}
\usepackage{float}
\usepackage[dvipsnames]{xcolor}
\usepackage{cite}

\DeclareMathOperator*{\argmax}{arg\,max}
\DeclareMathOperator*{\argmin}{arg\,min}
\newtheorem{definition}{Definition}

\newtheorem{assumption}{Assumption}

\usepackage{graphicx} 
\DeclareMathSymbol{\shortminus}{\mathbin}{AMSa}{"39}

\makeatletter
\def\adl@drawiv#1#2#3{%
        \hskip.5\tabcolsep
        \xleaders#3{#2.5\@tempdimb #1{1}#2.5\@tempdimb}%
                #2\z@ plus1fil minus1fil\relax
        \hskip.5\tabcolsep}
\newcommand{\cdashlinelr}[1]{%
  \noalign{\vskip\aboverulesep
           \global\let\@dashdrawstore\adl@draw
           \global\let\adl@draw\adl@drawiv}
  \cdashline{#1}
  \noalign{\global\let\adl@draw\@dashdrawstore
           \vskip\belowrulesep}}
\makeatother

\title{\LARGE \bf
Inverse Mixed Strategy Games with Generative Trajectory Models
}

\author{Max Muchen Sun, Pete Trautman, and Todd Murphey
\thanks{This work is supported by the Honda Research Institute Grant HRI-001479. The views expressed are those of the authors and do not necessarily reflect those of the funding institutions.}
\thanks{Max Muchen Sun and Todd Murphey are with the Department of Mechanical Engineering, Northwestern University, Evanston, IL 60208, USA. Pete Trautman is with Honda Research Institute, San Jose, CA 95134, USA. Email: {\tt\small msun@u.northwestern.edu}. Project website: \url{https://sites.google.com/view/inverse-mixed-strategy/}}
}

\begin{document}
\allowdisplaybreaks

\bstctlcite{IEEEexample:BSTcontrol}

\maketitle
\thispagestyle{empty}
\pagestyle{plain}

\begin{abstract}
Game-theoretic models are effective tools for modeling multi-agent interactions, especially when robots need to coordinate with humans. However, applying these models requires inferring their specifications from observed behaviors---a challenging task known as the inverse game problem. Existing inverse game approaches often struggle to account for behavioral uncertainty and measurement noise, and leverage both offline and online data. To address these limitations, we propose an inverse game method that integrates a generative trajectory model into a differentiable mixed-strategy game framework. By representing the mixed strategy with a conditional variational autoencoder (CVAE), our method can infer high-dimensional, multi-modal behavior distributions from noisy measurements while adapting in real-time to new observations. We extensively evaluate our method in a simulated navigation benchmark, where the observations are generated by an unknown game model. Despite the model mismatch, our method can infer Nash-optimal actions comparable to those of the ground-truth model and the oracle inverse game baseline, even in the presence of uncertain agent objectives and noisy measurements.
\end{abstract}

\IEEEpeerreviewmaketitle


\section{Introduction}

It is increasingly important for robots to interact with humans without causing discomfort or compromising task efficiency. Examples include social navigation~\cite{trautman_robot_2013,mavrogiannis_core_2023} and autonomous driving~\cite{schwarting_planning_2018,schwarting_social_2019}. In such scenarios, the robot must reason over the influence of its actions on others' decision-making. Such reasoning capability requires the conventional single-agent planning models to be replaced by multi-agent interaction models, enabling the robot to plan its actions while simultaneously predicting other agents' responses. 

Game theory provides powerful models for multi-agent interactions. The notion of Nash equilibrium~\cite{nash_non-cooperative_1951}, where each agent takes optimal actions that depend on other agents' optimal actions, is a rigorous yet flexible principle for multi-agent decision-making. However, applying game-theoretic models in practice requires accurately specifying agent objectives. Misalignment in these objectives can result in unsafe behavior, making the inverse game problem---inferring objectives from observed behaviors---vital for the effectiveness of such models.

\begin{figure}
    \centering
    \includegraphics[width=0.49\textwidth]{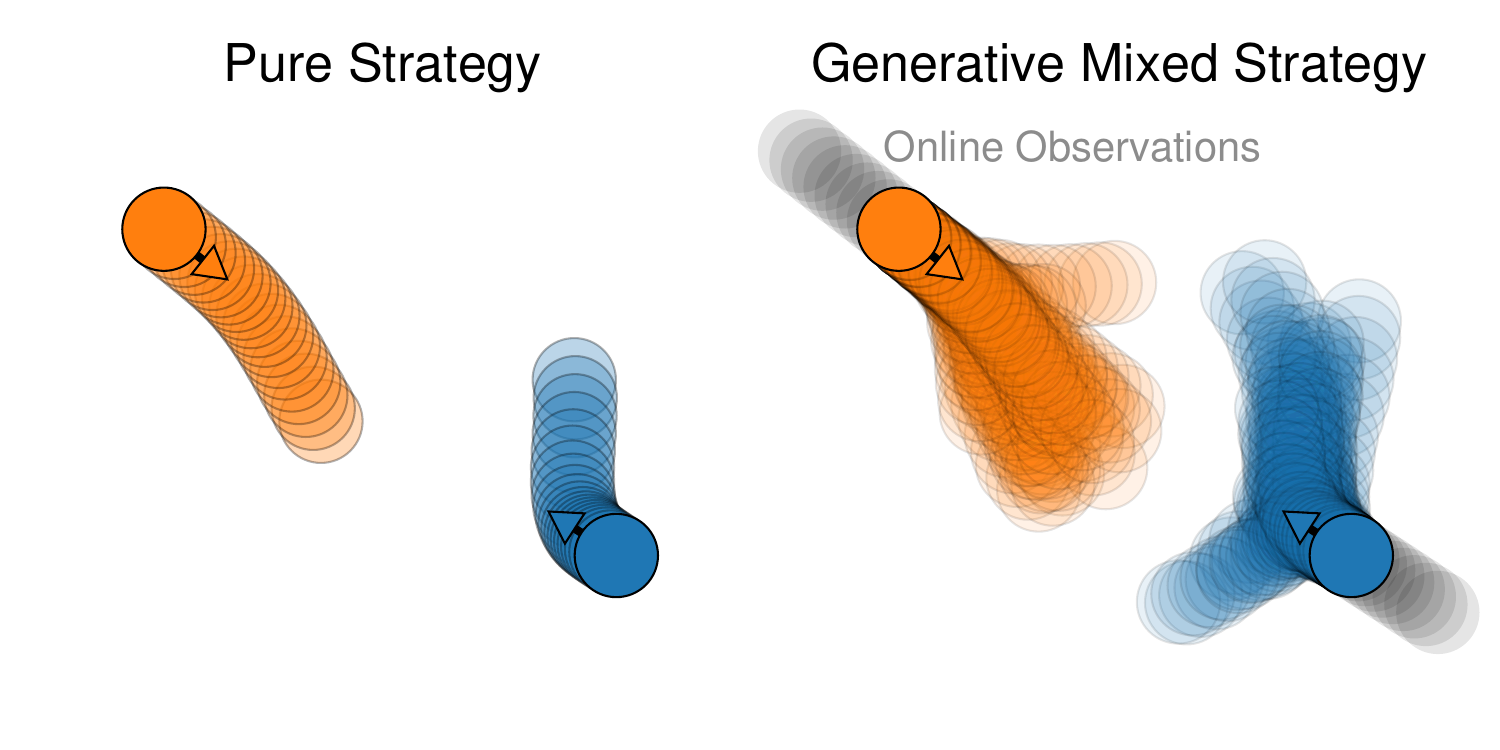}
    \vspace{-3em}
    \caption{The pure strategy inverse game formula (left) often fails to handle uncertainty or infer from both offline and online data, while our mixed strategy formula (right) addresses these issues by inferring multi-modal distributions with a generative trajectory model.}
    \label{fig:intro}
    \vspace{-1.5em}
\end{figure}

In this work, we propose an inverse game method based on a mixed strategy game model. Unlike commonly used pure strategy inverse game methods, which assume that observed agent actions stem from strictly optimal and deterministic decisions without measurement noise, the mixed strategy model treats observed actions as samples from action distributions, naturally accounting for uncertainty (as illustrated in Fig.~\ref{fig:intro}). We represent each agent's mixed strategy using a conditional variational autoencoder (CVAE)~\cite{sohn_learning_2015} and specify the agent objective as a neural network, the training of which takes less than 20 minutes with less than 100 demonstrations. By leveraging the CVAE-based generative trajectory model, our method infers high-dimensional, multi-modal behavior distributions from noisy offline data while adapting in real-time to new online observations. We evaluate our approach in a canonical task of resolving multiple paths intersecting at a single point, where a robot interacts with four other agents under an unknown game model with noisy measurements. Our results show that the method can infer actions comparable to those of the ground-truth model and the oracle inverse game baseline, both in terms of safety and runtime cost.

\section{Preliminaries and Related Works}

In this section, we simultaneously present the basics of game theory models, inverse games, and related work.

\subsection{Notations}
We assume $N$ agents indexed by the set $\mathcal{I}{=}\{1,{\dots},N\}$. The state of agent $i{\in}\mathcal{I}$ at time $t$ is denoted as $x_i^{t}$, following which we denote the trajectory of the agent from time $t_1$ to time $t_2$ as $x_i^{t_1:t_2}$. For both the state and trajectory notations, we use the subscript $\mathcal{I}$ to denote the set of all agents' states and trajectories, such that $x_{\mathcal{I}}^{t}{=}\{x_i^t\}_{i\in\mathcal{I}}$ and $x_{\mathcal{I}}^{t_1:t_2}{=}\{x_i^{t_1:t_2}\}_{i\in\mathcal{I}}$. We assume the states of all the agents are within the same state set $\mathcal{X}$. 

The temporal evolution of each agent's state is governed by the dynamics $x_i^{t+1}{=}f_i(x_i^t, u_i^t)$, where $u_i^t$ is the control. Under the dynamics constraints, we denote the set of feasible trajectories between time $t_1$ and $t_2$ of all agents as $\mathcal{S}^{t_1:t_2}$. 

A probability distribution of a trajectory between time $t_1$ and $t_2$ is denoted as $p(x_i^{t_1:t_2})$. We denote the set of all feasible probability density functions of trajectories between time $t_1$ and $t_2$ as $\mathcal{P}^{t_1:t_2}{=}\{ p \vert p{:}\mathcal{S}^{t_1:t_2}{\mapsto}\mathbb{R}_0^+, \int_{\mathcal{S}^{t_1:t_2}} p(s) ds {=} 1 \}$. Similar to the trajectory notations, we use the subscript $p_i$ to denote the probability distribution of agent $i$'s trajectory. 

\subsection{Games and Nash equilibrium}

As one of the pivotal works for applying game-theoretic models in interactive motion planning, a two-agent stochastic game model is introduced in~\cite{sadigh_planning_2016} for autonomous vehicles to coordinate with human drivers. Since then, efforts have been put into developing scalable and computationally efficient models~\cite{williams_best_2018,fridovich-keil_efficient_2020,lecleach_algames_2022} and accounting for uncertainty~\cite{wang_game-theoretic_2020,schwarting_stochastic_2021,mehr_maximum-entropy_2023,so_mpogames_2023,chen_soft-bellman_2023,lidard_blending_2024}.

In this work, we focus on trajectory games, where the decision of each agent is represented by their trajectory, but the problem formulation and the derivations are compatible with and can be generalized to other games. 

\begin{definition}[Pure strategy trajectory game]
    Given the time window $t_0$ to $t_f$, an N-player pure strategy trajectory game is defined as the tuple of $(\mathcal{I}, \mathcal{S}^{t_0:t_f}, \mathcal{J})$:
    \begin{align}
        \mathcal{J} = \left\{ J_i\left(x_1^{t_0:t_f}, {\dots}, x_i^{t_0:t_f}, {\dots}, x_N^{t_0:t_f}\right) \right\}_{i\in\mathcal{I}}, \label{eq:def_pure_game}
    \end{align} where $\mathcal{I}$ is the agent index set, $\mathcal{S}^{t_0:t_f}$ is the set of feasible trajectories within the time window, $\mathcal{J}$ is the set of trajectory objectives of all agents, and each agent's planned trajectory is called a pure strategy.
\end{definition}

\begin{definition}[Pure strategy Nash equilibrium]
    Given the game $(\mathcal{I}, \mathcal{S}^{t_0:t_f}, \mathcal{J})$, a pure strategy Nash equilibrium is a set of trajectories $\{s_i^* \in \mathcal{S}^{t_0:t_f} \}_{i\in\mathcal{I}}$ that satisfies the following:
    \begin{align}
        s_i^* = \argmin_{s_i} J_i\left( s_1^*, {\dots}, s_i, {\dots}, s_N^* \right), \text{ } \forall i \in \mathcal{I}. \label{eq:def_pure_nash}
    \end{align}
\end{definition}

\begin{definition}[Mixed strategy trajectory game] \label{def:mixed_game}
    Given the time window $t_0$ to $t_f$, an N-player mixed strategy trajectory game is defined as the tuple of $(\mathcal{I}, \mathcal{S}^{t_0:t_f}, \mathcal{P}^{t_0:t_f}, \mathcal{J})$, where the objective function set $\mathcal{J}$ is defined in (\ref{eq:def_pure_game}). 
\end{definition}

\begin{definition}[Mixed strategy Nash equilibrium]
    Given the game $(\mathcal{I}, \mathcal{S}^{t_0:t_f}, \mathcal{P}^{t_0:t_f}, \mathcal{J})$, a mixed strategy Nash equilibrium is a set of trajectory probability distributions $\{p_i^*\in\mathcal{P}^{t_0:t_f}\}_{i\in\mathcal{I}}$ that satisfies:
    \begin{align}
        p_i^* = \argmin_{p_i} \mathbb{E}_{p_1^*,{\dots},p_i,{\dots},p_N^*}\left[ J_i \right], \text{ } \forall i \in \mathcal{I}, \label{eq:def_mixed_nash}
    \end{align} where the trajectory distribution of each agent is called a mixed strategy, and $\mathbb{E}_{p_1,{\dots},p_N}[J]$ denotes joint expectation.
\end{definition}

Lastly, depending on the objective function specifications, there could exist more than one Nash equilibrium for both pure strategy and mixed strategy games. 

\subsection{Inverse games}

The inverse game problem is often framed as a parameter estimation problem for a pure strategy game, solved using maximum likelihood estimation MLE)~\cite{sadigh_planning_2016,mehr_maximum-entropy_2023,liu_learning_2023,li_cost_2023,peters_online_2023} or Bayesian inference~\cite{le_cleach_lucidgames_2021,schwarting_social_2019,liu_auto-encoding_2024}, with online or offline observations. Additionally, a related branch of work focuses on online inference of local Nash equilibrium from runtime observations, where the game model is predefined, but multiple locally optimal Nash equilibrium solutions may exist~\cite{peters_inference-based_2020,so_mpogames_2023,peters_contingency_2024}.

\begin{definition}[Inverse pure strategy game] \label{def:inverse_pure_game}
Given a multi-agent joint trajectory dataset $\mathcal{D}^{t_0:t_f}$ defined as:
\begin{align}
    \mathcal{D}^{t_0:t_f} = \{ (x_{1,d}^{t_0:t_f}, {\dots}, x_{N,d}^{t_0:t_f}) \}_d, \label{eq:inverse_game_dataset}
\end{align} and a parameterized pure strategy game $(\mathcal{I}, \mathcal{S}^{t_0:t_f}, \mathcal{J}_{\theta})$:
\begin{align}
    \mathcal{J}_{\theta} = \left\{ J_{i,\theta}\left(x_1^{t_0:t_f}, {\dots}, x_i^{t_0:t_f}, {\dots}, x_N^{t_0:t_f} \right) \right\}_{i\in\mathcal{I}}, \label{eq:inverse_game_obj}
\end{align} the inverse pure strategy game problem involves fitting the game model parameter $\theta$ to the joint trajectory dataset.
\end{definition}

\begin{assumption}[Pure strategy Nash-optimality] \label{assumption:nash_optimality}
    Observed behaviors in the dataset (\ref{eq:inverse_game_dataset}) are governed by a pure strategy Nash equilibrium (\ref{eq:def_pure_nash}) $\{s_{i,\theta}^* {\in} \mathcal{S}^{t_0:t_f} \}_{i\in\mathcal{I}}$ under an unknown game model parameter $\theta$. The joint trajectories in the dataset (\ref{eq:inverse_game_dataset}) are samples from a measurement model:
    \begin{align}
        x_{i,d}^{t_0:t_f} \sim g\left( x^{t_0:t_f} \vert s_{i,\theta}^* \right), \text{ } \forall i \in \mathcal{I}.
    \end{align}
\end{assumption}

Based on Assumption~\ref{assumption:nash_optimality}, the inverse game problem can be solved as a maximum likelihood estimation (MLE) problem:
\begin{gather}
    \theta^* = \argmax_{\theta} \prod_{d} \prod_{i=1}^{N} g\left( x_{i,d}^{t_0:t_f} \Big\vert s_{i,\theta}^* \right), \label{eq:mle_inverse_game} \\
    \text{s.t. } s_{i,\theta}^* = \argmin_{s_i} J_{i,\theta}\left( s_{1,\theta}^*, {\dots}, s_i, {\dots}, s_{N,\theta}^* \right), \text{ } \forall i \in \mathcal{I}. \nonumber
\end{gather} The MLE formula reduces to a least-squares estimation problem if the conditional distribution is Gaussian. MLE methods provide a point estimation $\theta^*$ of the pure strategy game parameter, which could lead to unsafe actions under uncertainty~\cite{kedia_game-theoretic_2023,liu_auto-encoding_2024}. An alternative approach is Bayesian inference, which infers a posterior distribution of the game parameter $\theta$, but it is often limited by the computation cost. In~\cite{liu_auto-encoding_2024}, a variational autoencoder (VAE) is trained on offline data to mitigate the computation cost of online Bayesian inference. However, the inference is limited to lower-dimensional parameters (e.g., navigation goal), and the pure strategy game formula used by the method does not inherently account for uncertainty in decision-making.

\subsection{Inverse mixed strategy games}

In this work, we take a different approach to integrate uncertainty into inverse games. We first extend Definition~\ref{def:inverse_pure_game} and Assumption~\ref{assumption:nash_optimality} from pure strategy to mixed strategy.

\begin{definition}[Inverse mixed strategy game] \label{def:inverse_mixed_game}
Given a joint trajectory dataset $\mathcal{D}^{t_0:t_f}$ (\ref{eq:inverse_game_dataset}) and a parameterized mixed strategy trajectory game $(\mathcal{I}, \mathcal{S}^{t_0:t_f}, \mathcal{P}^{t_0:t_f}, \mathcal{J}_{\theta})$, the inverse mixed strategy game problem involves inferring the game model parameter $\theta$ from the joint trajectory dataset.
\end{definition}

\begin{assumption}[Mixed strategy Nash-optimality] \label{assumption:mixed_nash_optimality}
    The behaviors in the dataset (\ref{eq:inverse_game_dataset}) are governed by a mixed strategy Nash equilibrium (\ref{eq:def_mixed_nash}) $\{p_{i,\theta}^*\in\mathcal{P}^{t_0:t_f}\}_{i\in\mathcal{I}}$ under an unknown game model parameter $\theta$. The joint trajectories in the dataset (\ref{eq:inverse_game_dataset}) are samples from the mixed strategy Nash equilibrium:
    \begin{gather}
        x_{i,d}^{t_0:t_f} \sim p_{i,\theta}^*\left( x^{t_0:t_f} \right), \text{ } \forall i \in \mathcal{I}. \label{eq:mixed_nash_assumption}
    \end{gather}
\end{assumption}

Based on Assumption~\ref{assumption:mixed_nash_optimality}, we can solve the inverse mixed strategy game problem as a MLE problem similar to (\ref{eq:mle_inverse_game}):
\begin{gather}
    \theta^* = \argmax_{\theta} \prod_{d} \prod_{i=1}^{N} p_{i,\theta}^*\left( x_{i,d}^{t_0:t_f} \right), \label{eq:mle_inverse_mixed} \\
    \text{s.t. } p_{i,\theta}^* = \argmin_{p_i} \mathbb{E}_{p_{1,\theta}^*,{\dots},p_i,{\dots},p_{N,\theta}^*}\left[ J_{i,\theta} \right], \text{ } \forall i \in \mathcal{I} \nonumber.
\end{gather} Although the MLE formula still provides a point estimate, unlike pure strategy inverse games, the mixed strategy game model naturally incorporates uncertainty by representing inferred decisions as probability distributions, thereby relaxing the strict rationality assumption in Assumption 1.

In~\cite{peters_learning_2022}, an inverse mixed strategy game method is proposed for both offline and online observations. However, its mixed strategy is limited to a fixed number of trajectory samples, which needs to be predetermined before learning. In the next section, we introduce our inverse mixed strategy game method, which uses generative trajectory models---conditional variational autoencoders (CVAEs)---as the mixed strategy for inference with both offline and online observations. Variational autoencoders (VAEs) and CVAEs have been applied for generative trajectory modeling in trajectory prediction~\cite{salzmann_trajectron_2020,ivanovic_multimodal_2021}, reinforcement learning~\cite{co-reyes_self-consistent_2018}, and self-supervised behavior analysis~\cite{sun_task_2021}.

\section{Inverse Mixed Strategy Games Using Generative Trajectory Models}

\subsection{Forward mixed strategy games}

In this work, we use the generalized mixed strategy games formula from~\cite{muchen_sun_mixed_2024}, which can be solved efficiently through an iterative algorithm named \emph{Bayesian recursive Nash equilibrium} (BRNE). We first give an overview of the game formula and the BRNE algorithm. 

The parameterized objective function of agent $i$, denoted as $J_{i,[\theta,\phi]}(x_1^{t_0:t_f},{\dots}, x_N^{t_0:t_f} )$, is specified as:
\begin{align}
    J_{i,[\theta,\phi]} {=} \sum_{j\in\mathcal{I}}^{j\neq i} l_{\theta}\left(x_i^{t_0:t_f}, x_j^{t_0:t_f}\right) {+} \log\left( \frac{p_i( x_i^{t_0:t_f} )}{q_{i,\phi}( x_i^{t_0:t_f} )} \right). \label{eq:mixed_traj_obj}
\end{align} In the above formula, $l_{\theta}(\cdot,\cdot)$ denotes a parameterized cost function that evaluates over two joint pure strategies (trajectories), $p_i(\cdot)$ is the mixed strategy of agent $i$. Furthermore, $q_{i,\phi}(\cdot)$ is a parameterized trajectory distribution we name as the \emph{nominal strategy} of agent $i$, representing agent $i$'s individual intent without the presence of other agents. We will discuss the details of $q_{i,\phi}$ in Section~\ref{subsection:cvae}. 

Following Definition~\ref{def:mixed_game} and based on results from~\cite{muchen_sun_mixed_2024} (Section IV, Eq.18), the mixed strategy Nash equilibrium (\ref{eq:def_mixed_nash}) $\{p_{i}^*\in\mathcal{P}^{t_0:t_f}\}_{i\in\mathcal{I}}$ for (\ref{eq:mixed_traj_obj}) is:
\begin{align}
    p_{i}^* = \argmin_{p_i} \sum_{j\in\mathcal{I}}^{j\neq i} \mathbb{E}_{p_i, p_{j}^*}[l_{\theta}] + D(p_i \Vert q_{i,\phi}), \forall i \in \mathcal{I}. \label{eq:mixed_prob_obj} 
\end{align} where $D(p\Vert q)$ is the Kullback-Leibler (KL) divergence.

There are two terms in (\ref{eq:mixed_prob_obj}): the first term captures the expected cost between agent $i$ and other agents, while the second term preserves agent $i$'s nominal mixed strategy. Note that the relative weight between the two terms can be a parameter of the cost function $l_{\theta}$. This structure of combining expected cost with a KL-divergence regulatory term is not unique in this specific game formula, with examples including model predictive path integral (MPPI) control~\cite{theodorou_generalized_2010,theodorou_relative_2012} and the corresponding game formula~\cite{williams_best_2018}. Furthermore, a similar game formula introduced in~\cite{lidard_blending_2024} views the nominal mixed strategy $q_{i,\phi}$ as a priori behavioral distribution that can be learned from data. We will take a similar approach to specify the nominal mixed strategy as a generative trajectory model in Section~\ref{subsection:cvae}. 

To solve the mixed strategy game (\ref{eq:mixed_prob_obj}), the Bayesian recursive Nash equilibrium (BRNE) algorithm is proposed in~\cite{muchen_sun_mixed_2024}, which iteratively updates the mixed strategies of all agents in a sequential order. Denote the current iteration number as $m$ and the mixed strategy of agent $i$ at the current iteration as $p_i^{[m]}(x_i^{t_0:t_f})$, the algorithm sequentially updates each agent's mixed strategy through the following:
\begin{align}
    & p_i^{[m+1]} = \eta \cdot p_i^{[m]}\left( x_i^{t_0:t_f} \right) \cdot \exp\left( -h^{[m]}_i\left(x_i^{t_0:t_f}\right) \right), \label{eq:brne_update_prob} \\
    & h^{[m]}_i = \sum_{j\in\mathcal{I}}^{j<i} \mathbb{E}_{p_j^{[m+1]}}\left[ 
l_{\theta}( x_i^{t_0:t_f}, \cdot) \right] {+} \sum_{j\in\mathcal{I}}^{j>i} \mathbb{E}_{p_j^{[m]}}\left[ 
l_{\theta}( x_i^{t_0:t_f}, \cdot) \right], \nonumber
\end{align} where $\eta$ is the normalization term and $p_i^{[0]}{=}q_{i,\phi}$. However, directly evaluating (\ref{eq:brne_update_prob}) is intractable. Instead, the mixed strategy $p_i^{[m]}(x_i^{t_0:t_f})$ is represented as $K$ weighted samples $\{ (x_{i,k}^{t_0:t_f}, w_{i,k}^{[m]}) \}_{k}$, where the samples are initially generated from the nominal mixed strategy $x_{i,k}^{t_0:t_f}{\sim}q_{i,\phi}(x_{i}^{t_0:t_f})$, such that (\ref{eq:brne_update_prob}) can be asymptotically approximated as:
\begin{align}
    w_{i,k}^{[m+1]} & = \eta \cdot w_k^{[m]} \cdot \exp\left( -\bar{h}^{[m]}_i\left(x_{i,k}^{t_0:t_f}\right) \right), \label{eq:brne_update_sample} \\
    \bar{h}^{[m]}_i\left(x_{i,k}^{t_0:t_f}\right) & = \sum_{j\in\mathcal{I}}^{j<i} \sum_{k^\prime=1}^{K} w_{j,k}^{[m+1]} \cdot l_{\theta}\left( x_{i,k}^{t_0:t_f}, x_{j,k^\prime}^{t_0:t_f} \right) \nonumber \\
    & \quad + \sum_{j\in\mathcal{I}}^{j>i} \sum_{k^\prime=1}^{K} w_{j,k}^{[m]} \cdot l_{\theta}\left( x_{i,k}^{t_0:t_f}, x_{j,k^\prime}^{t_0:t_f} \right),
\end{align} where $\eta$ is the normalization term to ensure $\sum_{k} w_{i,k}^{[m+1]} {=} 1$. The full iterative process of BRNE is described in Algorithm~\ref{algo:forward_game}. BRNE guarantees the convergence to a Nash equilibrium and a lower-bounded reduction of the joint expected cost between agents, we refer the readers to~\cite{muchen_sun_mixed_2024} for details on the formal properties of Algorithm~\ref{algo:forward_game}. As we can see in (\ref{eq:brne_update_sample}) and Algorithm~\ref{algo:forward_game}, given the initial samples from the nominal mixed strategies, the calculations involved in weight update are fully differentiable, which is a crucial property for us to solve the inverse game in (\ref{eq:mle_inverse_mixed}). 

\begin{algorithm} [t!]
    \caption{Forward mixed strategy game (\ref{eq:mixed_prob_obj})}
    \label{algo:forward_game}
    \begin{algorithmic}[1] 
        \Procedure{ForwardGame}{$l_\theta, \{x_{i,k}^{t_0:t_f}\}_{i\in\mathcal{I},k}$}
        \State $m \gets 0$ \Comment{$m$ is the iteration index.}
        \For{each sample $k$ of each agent $i\in\mathcal{I}$}
            \State Initialize weights $\{w_{i,k}^{[m]}\}_k \gets \{\frac{1}{K}\}_k$
        \EndFor
        \While{exit criterion not met}
            \For{each agent $i\in\mathcal{I}$}
                \State Update weight $w_{i,k}^{[m+1]}$ of each sample  (\ref{eq:brne_update_sample})
            \EndFor
            \State $m \gets m+1$
        \EndWhile
        \State \textbf{return} $\{w_{i,k}^{[m]}\}_{i\in\mathcal{I},k}$
        \EndProcedure
    \end{algorithmic}
\end{algorithm}

\begin{algorithm} [t!]
    \caption{Inverse mixed strategy game (one iteration)}
    \label{algo:inverse_game}
    \begin{algorithmic}[1] 
        \Procedure{InverseGame}{$\theta, \{\zeta_d^t\}_{t,d}$} \Comment{$\zeta_d^t$ in (\ref{eq:tuple_data_point})}
        \State $loss \gets 0$ \Comment{Initialize loss} 
        \For{Each $\zeta_d^t$ in the batch} 
            \State Generate nominal samples $\{ s_{i,k,d}^t \}_{i,k}$ from (\ref{eq:full_nominal_strategy})
            \State $\{w_{i,k,d}^{t}\}_{i,k} {\gets}$ \textsc{ForwardGame}$(l_{\theta}, \{ s_{i,k,d}^t \}_{i,k})$
            \For{Each agent $i$}
                \State $loss \gets loss + \sum_{k} w_{i,k,d}^{t} \cdot \left\Vert \bar{s}_{i,d}^{t} - s_{i,k,d}^t  \right\Vert^2$ 
            \EndFor 
        \EndFor
        \State $\theta \gets \textsc{BackPropagation}(\theta, loss)$
        \State \textbf{return} $\theta$
        \EndProcedure
    \end{algorithmic}
\end{algorithm}

To solve the inverse game problem for the formula (\ref{eq:mixed_prob_obj}), we optimize the parameter $\theta$ and $\phi$ separately to make the computation tractable. In the next section, we will first discuss constructing the nominal mixed strategies $q_{i,\phi}$ using generative trajectory models. 

\subsection{Generative trajectory models as mixed strategies} \label{subsection:cvae}

Denote an \emph{offline} dataset of multi-agent joint trajectories between time $0$ to $T$ as $\mathcal{D}^{0:T} {=} \{(x_{1,d}^{0:T}, \dots, x_{N,d}^{0:T})\}_d$, we specify the nominal mixed strategy as a parameterized conditional probability distribution from a receding horizon perspective:
\begin{align}
    q_{i,\phi}( x_i^{t: t+\tau_1} ) {=} p_{\phi}( x_i^{t: t+\tau_1} \big\vert x_i^{t\shortminus\tau_2: t}, \beta_i ), \forall t \in [\tau_2, T{\shortminus}\tau_1], \label{eq:cond_prob}
\end{align} where $\tau_1$ is the horizon of inferred future states, $\tau_2$ is the horizon of observed past states, $\beta_i$ is a task-relevant variable of agent $i$ given a priori (e.g., navigation goal), and the parameter $\phi$ characterizes the conditional distribution. We can optimize the parameter $\phi$ by solving the following maximum likelihood estimation problem:
\begin{align}
    \phi^* = \argmax_{\phi} \prod_{d} \prod_{i=1}^{N} \prod_{t=\tau_2}^{T{\shortminus}\tau_1} p_{\phi}\left( x_{i,d}^{t: t+\tau_1} \Big\vert x_{i,d}^{t\shortminus\tau_2: t}, \beta_i \right). \label{eq:generative_mle}
\end{align} Note that even though we are given multi-agent joint trajectory data, the training of the nominal mixed strategy is based on the individual trajectory of each agent. In other words, the nominal mixed strategy does not capture any inter-agent interaction during training. 

In this work, we specify the conditional probability distribution (\ref{eq:cond_prob}) as a conditional variational autoencoder (CVAE). To simplify the notation, we replace $x_i^{t:t+\tau_1}$ with $s_i^{t}$ to denote the inferred future trajectory, and use $c_i^{t}=[x_i^{t\shortminus\tau_2:t}, \beta_i]$ to denote the joint conditional variable. The CVAE model represents the conditional probability distribution (\ref{eq:cond_prob}) as a marginalized distribution over a low-dimensional latent variable $z$ whose prior distribution is the standard Gaussian distribution. During the training, the conditional probability distribution (\ref{eq:cond_prob}) is approximated as:
\begin{align}
    p_{\phi}(s_i^t \vert c_i^t) \approx \int p_{o,\phi}(s_i^t \vert c_i^t, z) p_{e,\phi}(z \vert s_i^t, c_i^t) dz, \label{eq:cvae_training_prob}
\end{align} where $p_{e,\phi}$ is a probability distribution characterized by an \emph{encoder} network and $p_{o,\phi}$ is characterized by a \emph{decoder network}. After the training, we can only use the optimized decoder network $p_{o,\phi^*}$ for inference, in which case the nominal mixed strategy (\ref{eq:cond_prob}) is represented as:
\begin{align}
    q_i\left( s_i^t \right) = \int p_{o,\phi^*}(s_i^t \vert c_i^t, z) p_{\mathcal{N}}(z; 0, 1) dz, \label{eq:cvae_inference}
\end{align} From (\ref{eq:cvae_inference}), we can generate trajectory samples for Algorithm~\ref{algo:forward_game}. We refer the readers to~\cite{kingma_auto-encoding_2013,sohn_learning_2015} for details about training and inference with CVAE models. 

Using a CVAE to model the nominal mixed strategy (\ref{eq:cond_prob}) allows us to infer with both offline and online observations. The training of CVAE uses the offline dataset to capture the prior statistical pattern in agent behavior, and we can generate new nominal mixed strategies conditioned on online observations using the trained model. Next, we solve the overall inverse mixed strategy problem in (\ref{eq:mle_inverse_mixed}) by learning the cost function $l_{\theta}$ in (\ref{eq:mixed_traj_obj}).

\subsection{Inverse Mixed Strategy Game}

We model the inter-agent cost function $l_{\theta}(\cdot,\cdot)$ in (\ref{eq:mixed_traj_obj}) as a neural network and optimize the parameter $\theta$ over the same offline dataset $\mathcal{D}^{0:T} = \{(x_{1,d}^{0:T}, \dots, x_{N,d}^{0:T})\}_d$ as in (\ref{eq:generative_mle}). Similar to the training of the CVAE model, we take the receding horizon approach to separate each set of joint trajectories $(x_{1,d}^{0:T}, {\dots}, x_{N,d}^{0:T})$ into past observations and future states $\mathcal{D} = \{\zeta_d^{t}\}$:
\begin{align}
    \zeta_d^t {=} \left( (o_{1,d}^{t} {,\dots,} o_{N,d}^{t}){,} (\bar{s}_{1,d}^{t} {,\dots,} \bar{s}_{N,d}^{t}) \right).  \label{eq:tuple_data_point}
\end{align} Here we simplify the notation by denoting past states $x_{i,d}^{t\shortminus\tau_2:t}$ as $o_{i,d}^{t}$ and future states $x_{i,d}^{t:t+\tau_1}$ as $\bar{s}_{i,d}^{t}$ from the dataset. 

Following (\ref{eq:mixed_traj_obj}) and (\ref{eq:mixed_prob_obj}), we can specify the MLE problem introduced in (\ref{eq:mle_inverse_mixed}) for the inverse mixed strategy game with the log-likelihood formula below:
\begin{align}
    \theta^* & = \argmax_{\theta} \sum_{d} \sum_{i=1}^{N} \sum_{t=\tau_2}^{T-\tau_1} \log p_{i,d,\theta}^{t*}\left( \bar{s}_{i,d}^{t} \right). \label{eq:full_inverse_mixed_mle} 
\end{align} Evaluating $p_{i,d,\theta}^{t*}(\bar{s}_{i,d}^{t})$ requires solving the forward game
\begin{gather}
    p_{i,d,\theta}^{t*} {=} \argmin_{p_{i,d}^{t}} \sum_{j\in\mathcal{I}}^{j\neq i} \mathbb{E}_{p_{i,d}^t, p_{j,d,\theta}^{t*}}[l_{\theta}] {+} D( p_{i,d}^{t} \Vert q_{i,d,\phi^*}^{t} ), \label{eq:full_mixed_forward_game} \\
    q_{i,d,\phi^*}^{t}(s_i^t) = \int p_{o,\phi^*}(s_i^t \vert o_{i,d}^t, \beta_i, z) p_{\mathcal{N}}(z; 0, 1) dz. \label{eq:full_nominal_strategy}
\end{gather}

To evaluate the MLE objective (\ref{eq:full_inverse_mixed_mle}), Algorithm~\ref{algo:forward_game} serves as a differentiable solver for the forward game (\ref{eq:full_mixed_forward_game}), which represents the nominal mixed strategies as sets of trajectory samples: $\{ s_{i,k,d}^t \}_{k} \sim q_{i,d}^{t}(s_i^t)$.
The algorithm then returns optimal sample weights $\{w_{i,k,d}^{t}\}_{i\in\mathcal{I},k}$ such that the weighted samples represent the optimal mixed strategy from the corresponding mixed strategy Nash equilibrium (\ref{eq:full_mixed_forward_game}): $\{(s_{i,k,d}^t, w_{i,k,d}^{t})\}_{k} \sim p_{i,d,\theta}^{t*}(s_{i,d}^t).$ The sample-based mixed strategy representation approximates the log-likelihood (\ref{eq:full_inverse_mixed_mle}):
\begin{align}
    \log p_{i,d,\theta}^{t*}\left( \bar{s}_{i,d}^{t} \right) \approx -\sum_{k} w_{i,k,d}^{t} \cdot \left\Vert \bar{s}_{i,d}^{t} - s_{i,k,d}^t  \right\Vert^2.
\end{align} 

\begin{figure}[t]
    \centering
    \includegraphics[width=0.49\textwidth]{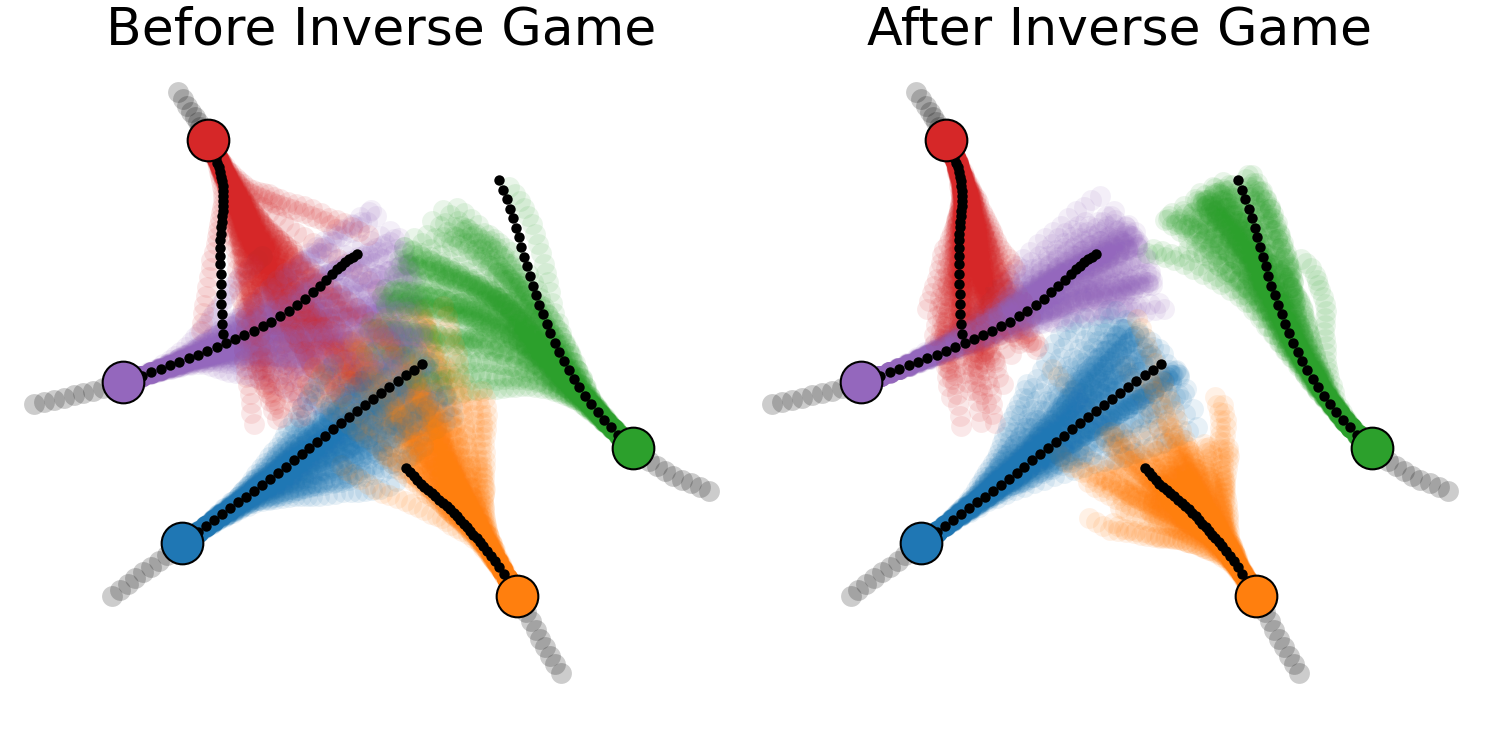}
    \vspace{-2.5em}
    \caption{The Nash-optimal mixed strategies before (left) and after (right) solving the inverse game (\ref{eq:full_inverse_mixed_mle}). Solid black lines represent the demonstrated future states in the dataset.}
    \label{fig:inverse_game}
    \vspace{-1em}
\end{figure}

Since the whole process of evaluating the MLE objective (\ref{eq:full_inverse_mixed_mle}) is differentiable, we can iteratively optimize the cost function parameter $\theta$ through backpropagation. We describe the process of one optimization iteration for the inverse mixed strategy game in Algorithm~\ref{algo:inverse_game}. Fig.~\ref{fig:inverse_game} shows an example of Nash equilibrium mixed strategies before and after solving the inverse game (\ref{eq:full_inverse_mixed_mle}).

\section{Evaluation}

\subsection{Benchmark design}

\noindent\textbf{[Simulation design] } We evaluate our inverse mixed strategy game method in a simulated navigation benchmark, where a group of 5 agents (one of them being the robot during tests) coordinate collision avoidance while reaching their respective destinations. The agents are simulated using the iLQGames algorithm~\cite{fridovich-keil_efficient_2020}, one of the most commonly used dynamic game solvers. The iLQGames algorithm calculates a local pure strategy Nash equilibrium, and we implement the algorithm as a model predictive controller (MPC) for each simulated agent except the robot. Each agent is modeled as a circular disk, and the state of each agent at time $t$ is denoted as $x_i^t{=}[\xi_{i,x}^t, \xi_{i,y}^t, \psi_i^t, v_i^t]$, which are x, y position, angle, and longitudinal velocity, respectively. The control is denoted as $u_i^t{=}[\omega_i^t,a_i^t]$, which are angular velocity and longitudinal acceleration. The dynamics of each agent is:
\begin{align}
    \dot{\xi}_{i,x}^t = v_i^t \cos(\psi_i^t), \text{ } \dot{\xi}_{i,y}^t = v_i^t \sin(\psi_i^t), \text{ } \dot{\psi}_i^t = \omega_i^t, \text{ } \dot{v}_i^t = a_i^t. \nonumber
\end{align} Each agent is assigned a navigation goal $g_i{=}[g_{i,x},g_{i,y}]$, a preferred longitudinal velocity $\bar{v}_i=1m/s$, and a reference trajectory $\bar{\xi}_i^t {=} [\bar{\xi}_{i,x}^t,\bar{\xi}_{i,y}^t]$, which is a straight line from the current position to the goal with the preferred longitudinal velocity. All agents are constrained with a minimal velocity $v_{min}{=}0.5m/s$, a maximal velocity $v_{max}{=}1.0m/s$, and a social zone distance of $\bar{d}{=}0.7m$ from other agents. The cost function of each agent consists of:
\begin{align}
    & \text{Navigation: } l_r(x_i^t) = (\xi_{i,x}^t{-}\bar{\xi}_{i,x}^t)^2 + (\xi_{i,y}^t{-}\bar{\xi}_{i,y}^t)^2, \nonumber \\
    & \text{Safety: } l_s(x_1^t{,\dots,}x_N^t) = \sum_{j\neq i} \mathbf{1}\{d_{i,j}{<}d_{min}\} {\cdot} (d_{i,j} {-} d_{min})^2, \nonumber 
\end{align} where $\mathbf{1}\{d_{i,j}{<}\bar{d}\}$ is an indicator function that is 1 if the distance $d_{ij}$ between agent $i$ and $j$ is smaller than $\bar{d}$ and 0 otherwise. The above two terms are combined as the cost function $l_i(x_1^t{,\dots,}x_N^t)$ for agent $i$ as:
\begin{align}
    l_i = 100 {\cdot} \lambda_i {\cdot} l_r(x_i^t) + 1000 {\cdot} (1 {\shortminus} \lambda_i) {\cdot} l_s(x_1^t{,\dots,}x_N^t) \label{eq:ilqgames_cost}, 
\end{align} where $\lambda_i\in[0,1]$ is an agent-specific cost parameter, and it will be randomly sampled during each test. \emph{Note that our method has no access to the agent cost functions and has no knowledge of the iLQGames algorithm.} 

\begin{figure}[t]
    \centering
    \vspace{-1em}
    \includegraphics[width=0.49\textwidth]{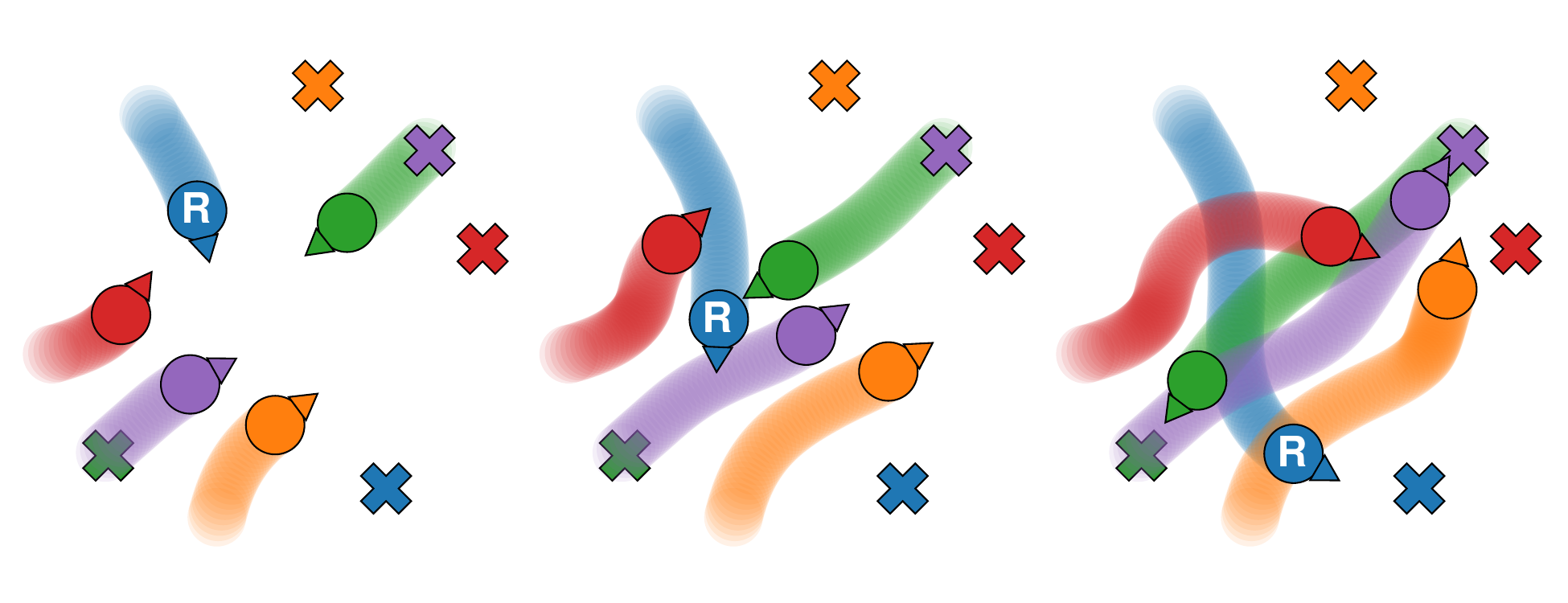}
    \vspace{-2em}
    \caption{Snapshots from a test with our method (blue: robot).}
    \label{fig:test_trial}
    \vspace{-0.5em}
\end{figure}

\noindent\textbf{[Task design] } In each test, all five agents (four iLQGames agents and one robot) are randomly initialized on a circle with a radius of 4$m$, with the navigation goals being the opposite of the circle. This design ensures that the shortest path for each agent will lead to collisions with others, making coordination necessary. The four game-theoretic agents are individually controlled by the iLQGames algorithm. The game agents have access to each other's true cost function, and they assume the robot is also controlled by the iLQGames algorithm with an assumed cost function. The robot has no access to the cost functions of game agents and makes decisions solely based on the inferred mixed strategy from our method. The inverse mixed strategy game is solved based on a dataset of navigation trials where all five agents are controlled by the iLQGames algorithm. Fig.~\ref{fig:test_trial} shows snapshots from a test trial of our method.

\noindent\textbf{[Data collection] } We collect 50 navigation trials as the training data. In each trial, we uniformly sample the initial position of each agent on the circle and uniformly sample the cost parameter $\lambda_i$ in (\ref{eq:ilqgames_cost}) between $0.1$ and $0.9$ for each agent, such that each navigation trial is governed by a different Nash equilibrium. The training data has an average length of 108 time steps per navigation trial. For training and testing with noisy observations, we add noise from a zero-mean isotropic Gaussian distribution with a standard deviation of $0.05$ and $0.1$ to the x and y positions of each agent.

\begin{figure*}[htbp]
    \centering
    \begin{subfigure}{0.49\textwidth}
        \centering
        \includegraphics[width=\linewidth]{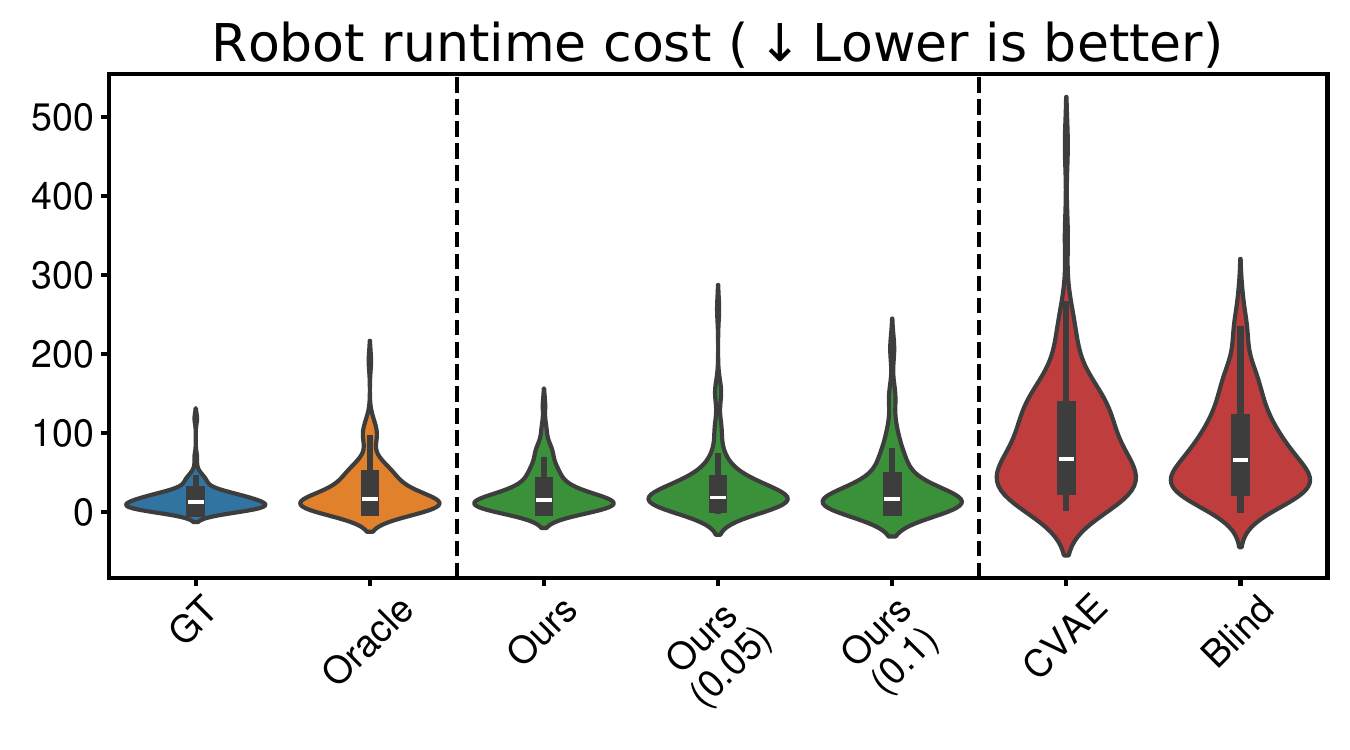} 
        \label{fig:robot_cost}
    \end{subfigure}
    \begin{subfigure}{0.49\textwidth}
        \centering
        \includegraphics[width=\linewidth]{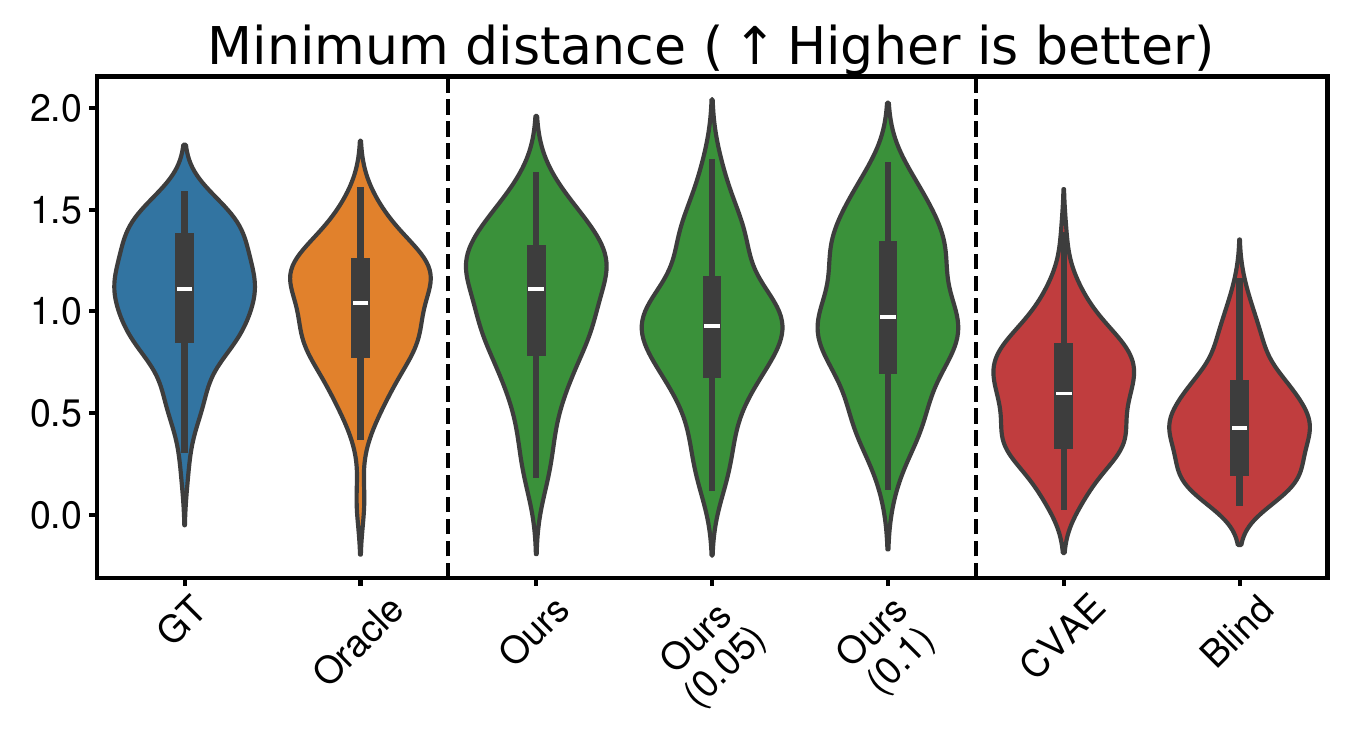}
        \label{fig:min_dist}
    \end{subfigure}
    \vspace{-1.8em}
    \caption{Quantitative results of the navigation benchmark. We show the median, quartiles, and distribution of the results. Our method has comparable performance with the ground-truth baseline and the oracle inverse game baseline, while not relying on privileged information (e.g., the iLQGames model and objective structure) and with noisy observations.}
    \vspace{-1.6em}
    \label{fig:benchmark_result}
\end{figure*}

\subsection{Implementation details}

We implement both the iLQGames algorithm and our method in JAX~\cite{bradbury_jax_2018} and Flax~\cite{heek_flax_2024}. The numbers of time steps for future states and past observations are $30$ and $10$ ($\tau_1$ and $\tau_2$ in (\ref{eq:cond_prob}). We separate the 50 navigation trials into 16900 pairs of past and future states through a sliding window approach for training. For the CVAE model, we use a gated recurrent unit (GRU) layer and 3 hidden layers with 256 dimensions as the encoder network, the same but mirrored architecture for the decoder network, and a 4-dimensional latent space. For the cost function $l_{\theta}$ in (\ref{eq:full_mixed_forward_game}), we use a multilayer perceptron (MLP) with 3 hidden layers of 256 dimensions. We train the CVAE for 50 epochs with a batch size of 32, and we train the MLP for 10 epochs with a batch size of 48, both with a learning rate of $1e{\shortminus}4$. The training of CVAE takes \emph{4 minutes} wall-clock time, and the training of MLP takes \emph{14 minutes} on an Nvidia RTX 6000. We use 100 trajectory samples per agent for training, and 200 samples per agent for inference. When using the CVAE model for motion planning, we use a moving waypoint toward the goal as a conditional variable to generate goal-reaching actions, and the inferred mixed strategy is converted to robot controls by solving a trajectory optimization problem to track the mean of the mixed strategy. The code of our implementation is available at \url{https://sites.google.com/view/inverse-mixed-strategy/}. 

\subsection{Baseline and metric selection}

\noindent\textbf{[Baselines] } There are three variants of our method: a noise-free variant with perfect measurement of other agents' states (\textsf{Ours}), and two variants with measurement noise from a zero-mean Gaussian distributions with standard deviation of $0.05$ (\textsf{Ours(0.05)}) and $0.1$ (\textsf{Ours(0.1)}). We compare our method with four baselines:

\textsf{GT}: The ground-truth baseline with access to all agents' cost functions and solves the exact iLQGames problem. 

\textsf{Oracle}: This method uses oracle maximum a posteriori estimation for the offline pure strategy inverse game problem, assuming knowledge of the iLQGames structure. Since the training data is generated with agent cost parameters $\lambda_i$ uniformly sampled between 0.1 and 0.9, the oracle posterior reflects this distribution. During testing, the method samples cost parameters from this posterior and solves the corresponding iLQGames problem for prediction and planning.

\textsf{CVAE}: This baseline only uses the CVAE model as the nominal mixed strategy without solving the mixed strategy game. It learns individual behavior from the training data without taking into account inter-agent relations.

\textsf{Blind}: This baseline solves the iLQGames problem with $\lambda_i{=}1$, such that the agent only optimizes the navigation cost and ignores the inter-agent safety cost.

We exclude online inverse game baselines due to the lack of accessible, intuitive implementations that can be integrated into our benchmark (e.g., those requiring CPU multi-threading). Additionally, implementing online inverse game for iLQGames falls outside the scope of this work.

\noindent\textbf{[Metrics] } We evaluate how closely the robot's actions align with the expectations of the other game agents by measuring the iLQGames runtime cost function assumed by the four agents for the robot, even though the robot does not make decisions based on this cost function. We evaluate the safety by measuring the minimum distance between the robot and other agents in each test. In addition, the collision rate is evaluated using the minimum distance between iLQGames agents as the threshold. We also report the numerical efficiency.

\subsection{Evaluation results}

Fig.~\ref{fig:benchmark_result} shows that our method performs comparably with the ground-truth model and oracle inverse game baseline, and significantly outperforms \textsf{CVAE} and \textsf{Blind} baselines in both robot runtime cost and minimum distance to other agents. The oracle inverse game baseline benefits from privileged knowledge of the iLQGames algorithm and cost function structure, which our method does not have access to. In real-world applications to human behavior data, such privileged information becomes assumptions that are unlikely to hold. The comparable performance of our method, despite lacking these assumptions, is highly encouraging. Collision rates are as follows: $0.0\%$(\textsf{GT}), $2.0\%$(\textsf{Oracle}), $1.0\%$(\textsf{Ours}), $2.0\%$(\textsf{Ours(0.05)}), $2.0\%$(\textsf{Ours(0.1)}), $8.0\%$(\textsf{CVAE}), and $17.0\%$(\textsf{Blind}). Our method performs consistently across varying observation noise and is faster, taking $0.08$ seconds per time step compared to $0.20$ seconds for iLQGames.

\section{Conclusion and Discussion}

In this work, we propose an inverse game method that leverages a generative trajectory model as the mixed strategy. Our simulations demonstrate that the method performs comparably to the ground truth model and the oracle baseline, even under uncertain agent objectives and observations. While this study focuses on a fixed number of homogeneous agents, our approach is not theoretically limited to such settings. Future work will extend the method to handle inhomogeneous agents and varying agent counts, as well as adapting other generative trajectory models such as diffusion-based models~\cite{chi_diffusion_2024}. Additionally, we plan to evaluate the method on unstructured real-world human trajectory datasets and conduct hardware experiments.

\bibliographystyle{IEEEtran}
\bibliography{references}

\end{document}